\documentclass[lettersize,journal]{IEEEtran}
\usepackage{amsmath,amsfonts}
\usepackage{algorithmic}
\usepackage{algorithm}
\usepackage{array}
\usepackage[caption=false,font=normalsize,labelfont=sf,textfont=sf]{subfig}
\usepackage{textcomp}
\usepackage{stfloats}
\usepackage{url}
\usepackage{verbatim}
\usepackage{graphicx}
\usepackage{cite}

\usepackage[pagebackref=true,breaklinks=true,letterpaper=true,colorlinks,bookmarks=false]{hyperref}

\usepackage{multirow}
\usepackage{times}
\usepackage{soul}
\usepackage{indentfirst}
\usepackage{booktabs}
\usepackage{threeparttable}

\begin{document}

\title{Controllable Person Image Synthesis with Spatially-Adaptive Warped Normalization}

\author{Jichao~Zhang,
        Aliaksandr~Siarohin,
        Hao~Tang, 
        Enver~Sangineto, \\
        Wei~Wang,
        Humphrey~Shi,
        Nicu~Sebe, Senior Member, IEEE
}


\markboth{}%
{Shell \MakeLowercase{\textit{et al.}}: A Sample Article Using IEEEtran.cls for IEEE Journals}


\maketitle

\begin{abstract}
Controllable person image generation aims to produce realistic human images with desirable attributes such as a given pose, cloth textures, or hairstyles. However, the large spatial misalignment between source and target images makes the standard image-to-image translation architectures unsuitable for this task. Most state-of-the-art methods focus on alignment for global pose-transfer tasks. However, they fail to deal with region-specific texture-transfer tasks, especially for person images with complex textures. To solve this problem, we propose a novel Spatially-Adaptive Warped Normalization~(SAWN) which integrates a learned flow-field to warp modulation parameters. It allows us to efficiently align person spatially-adaptive styles with pose features.
Moreover, we propose a novel Self-Training Part Replacement (STPR) strategy to refine the model for the texture-transfer task, which improves the quality of the generated clothes and the preservation ability of non-target regions. Our experimental results on the widely used DeepFashion dataset demonstrate a significant improvement of the proposed method over the state-of-the-art methods on pose-transfer and texture-transfer tasks. The code is available at \url{https://github.com/zhangqianhui/Sawn}.
\end{abstract}
\begin{IEEEkeywords}
Generative Adversarial Networks, Person Image Generation, Controllable Image Generation.
\end{IEEEkeywords}

\section{Introduction}
\IEEEPARstart{P}{erson} image generation has attracted much attention in computer graphics and computer vision due to its usefulness in data augmentation for surveillance~\cite{qian2018pose}, fashion design~\cite{han2020design}, and virtual try-on~\cite{han2018viton}. Compared to other common image types, a human image has rich variations in poses, clothing, hairstyles, body shape, self-occlusions, and other factors. Because of this, it is not easy to synthesize human images in a new target pose or wearing new clothing without having a 3D textured model for this specific person as supervision. Initial progress has been made in the pose-guided person image generation (pose-transfer) task. In the pose-transfer task, we are given a source image and a target pose and we are asked to generate a person from the source image in the target pose. The pose-guided person generation task was proposed by~\cite{ma2017pose} and some improved models have been invented~\cite{zhu2019progressive,siarohin2018deformable}. Most of them follow a GAN architecture with an autoencoder-like generator that takes as input the target pose (usually represented by keypoints) and the person appearance, represented by an RGB image.

\begin{figure}[!t]\small
\begin{center}
\vspace{-0.2cm}
\includegraphics[width=1.0\linewidth]{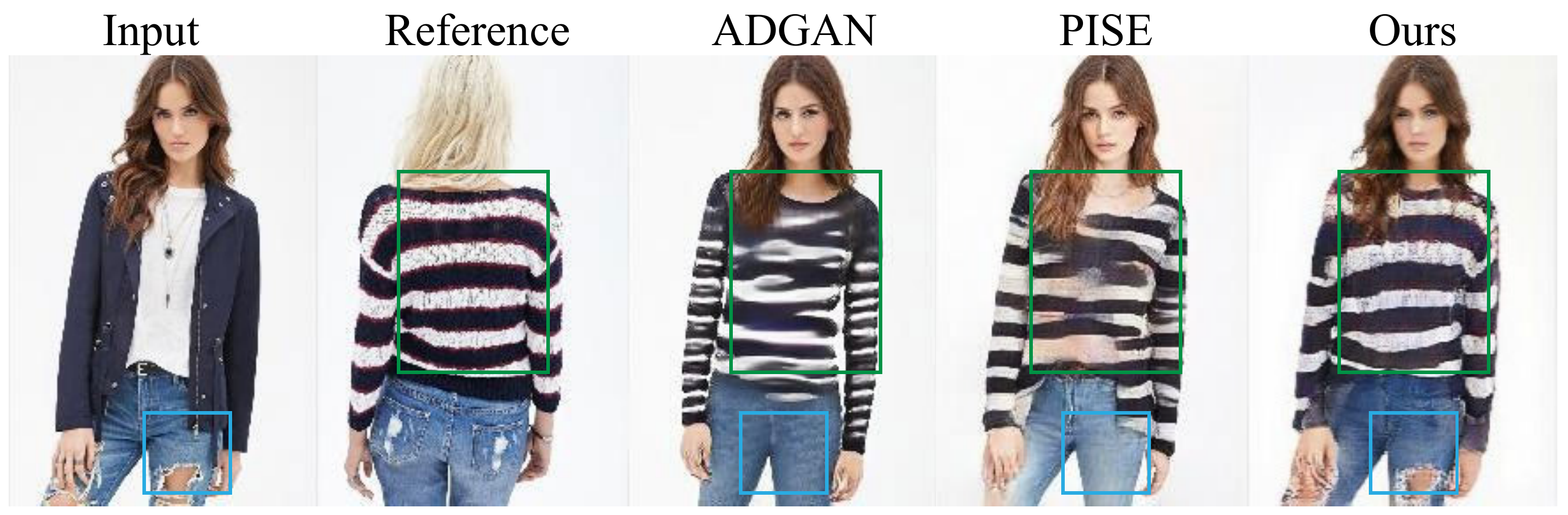}
\end{center}
\vspace{-0.3cm}
\caption{Comparison of texture transfer on clothes between our method, ADGAN~\cite{men2020controllable} and PISE~\cite{PISE}. Our model achieves better complex texture transfer~(green box) and better preserves non-target regions~(blue box).}
\label{fig:show}
\vspace{-0.4cm}
\end{figure}

Recently, ADGAN~\cite{men2020controllable} proposed an architecture for controllable person image generation, which can be regarded as a generalization of the pose-guided person image generation. ADGAN also allows the transfer of the texture of a particular person region while preserving the style of the non-target person regions. In more detail, given the source and reference images and the mask of a particular region, texture-transfer aims to copy the texture of the given segment from the reference image to the source image. Specifically, ADGAN is based on the `semantic image synthesis architecture'~\cite{wang2018pix2pixHD} with Adaptive Instance Normalization~(AdaIN)~\cite{Huang2017ArbitraryST}. However, as shown in Fig.~\ref{fig:show} (3rd column), ADGAN tends to synthesize very blurry textures which lack details. We argue that ADGAN learns the style parameters for AdaIN with multi-layer perceptrons (MLPs), which neglects the spatial information of styles in person appearance. For example, given a person image, the skin, garment, and hairstyles should have different style features. Even the style is spatially adaptive for different local regions of the garment. Thus, some researchers have proposed the spatially-adaptive normalization, which learns style parameters with spatial information for semantic image editing tasks~\cite{park2019SPADE,Zhu_2020_CVPR}. However, the spatially-adaptive instance normalization directly used for person image generation leads to the spatial misalignment problem in the feature space between style parameters and pose features. Recently, PISE~\cite{PISE} proposed a spatially-aware normalization and used VGG features to constrain pose and style features in the same domain. However, PISE also fails in generating and transferring complex textures~(Fig.~\ref{fig:show} (4th column)), as its constraint is very weak and lacks an explicitly spatial warping operation to align style parameters and pose features. 

To solve this problem, we propose a novel spatially-adaptive warped normalization~(SAWN) which takes the learned flow-field as a condition to warp the modulation parameters, scales, and bias in the normalization, to align the style features with the pose features. This operation is applied for multiple scales of the decoder. SAWN can solve the region-specific spatial misalignment problem by introducing the related mask of the person part for the texture-transfer task. We refer to SAWN with the mask as M-SAWN. Additionally, we propose a self-training part replacement strategy to refine the trained model. In more detail, we finetune the network by performing texture transfer via a person part replacement operation between the same person in different poses. This modification brings a significant improvement to the quality of the generated textures and the consistency of other regions (see Fig.~\ref{fig:show} (5th column)) with just an 100-step optimization.

In summary, the main contributions of this work are:
\begin{itemize}
  \item[1)] We propose a novel spatially-adaptive warped normalization (SAWN) integrating the learned flow field to warp the scales and bias parameters to align the style and pose features. Furthermore, we present M-SAWN to solve a region-specific spatial misalignment problem for the texture-transfer task.
  \item[2)] We propose a novel training strategy, i.e., the self-training part replacement (STPR), to refine the trained model reducing the discrepancy between the training and test phases, improving the quality of texture-transfer results, and attaining better preservation of some regions.
  \item[3)] Our method achieves state-of-the-art results on two challenging tasks, i.e., pose-transfer and texture-transfer, especially for complex textures.
\end{itemize}
\section{Related Work}
\noindent \textbf{Pose-Guided Person Image Generation.}
In the short span of five years, generative adversarial networks~(GANs)~\cite{goodfellow2014generative} opened the door to many creative applications and have come to dominate the field of image inpainting~\cite{iizuka2017globally}, image editing~\cite{tan2020michigan,portenier2018faceshop,abdal2020styleflow}, 2D and 3D image synthesis~\cite{zhu2017unpaired,zhang2018sparsely,choi2018stargan,han2018viton,abdal2019image2stylegan,park2019semantic,menon2020pulse,huang2018multimodal,alldieck2019tex2shape,Texler20-SIG} and person image generation~\cite{zhang20213daware,fruhstuck2022insetgan,Sarkar2021HumanGANAG}. Most models for person image generation are based on GANs and can be divided into two groups depending on the underlined target pose representation, i.e., 2D keypoints and SMPL correspondences~\cite{SMPL:2015} which are estimated using DensePose~\cite{guler2018densepose}.

The first group~\cite{ma2017pose,li2019dense,tang2020bipartite,ma2018disentangled,huang2020generating,Balakrishnan2018SynthesizingIO,tang2020xinggan,ren2022neural} exploits keypoints encoded as heatmaps. This approach was initially proposed by PG2~\cite{ma2017pose} which consists of two key stages: first, a U-Net is used to generate a coarse person image with a target pose, then another U-net is used to refine the image generated previously. However, the photorealistic person generation with accurate preservation of complex textures is beyond the reach of this model, as it cannot handle large spatial variations and non-rigid deformation between different poses. Other studies explored ways to handle the deformation between different poses~\cite{siarohin2018deformable,siarohin2021PAMI,zhu2019progressive,ren2020deep}. For instance, Siarohin et al.~\cite{siarohin2018deformable,siarohin2021PAMI} propose a deformable skip connection for the generator, which `moves' local information according to the structural deformations from different poses. However, this method requires pre-defined transformation components, which limits its applications. Zhu et al.~\cite{zhu2019progressive} propose progressive attention transfer blocks~(PATBs) at the feature level, which could focus on the local transfer in the manifold, therefore circumventing the difficulty of capturing the pose variation in the global structure. Then, Tang et al.~\cite{tang2020xinggan} optimize the PATBs module of Pose-Attn~\cite{zhu2019progressive} with a mutually learning module between appearance and pose. However, both methods still struggle to generate photorealistic person images with accurate preservation of the complex textures, as these methods do not explicitly learn the spatial transformation between different poses. Recently, some methods~\cite{tang2021structure,li2019dense,ren2020deep,lv2021learning} presented the warped module for person alignment with the pre-trained flow field, which has achieved very high-quality results for the pose-transfer generation. However, they cannot perform texture-transfer tasks as they lack the disentangled module for appearance. In contrast, our SAWN can solve the region-specific spatial misalignment problem for the texture-transfer task by introducing the related mask.


The second group~\cite{Sarkar2021HumanGANAG,neverova2018dense,Grigorev2018CoordinatebasedTI,Sarkar2020} is based on the rich 3D SMPL~\cite{SMPL:2015} correspondences as pose presentation. For instance, Neverova et al.~\cite{neverova2018dense} use DensPose to guide the proposed predictive module and the warping module. The warping module aims to warp the texture to the UV coordinates and performs the inpainting for UV texture, then warps back to the target image. At the same time, the predictive module is a generative model conditioned on DensePose. The results from the two modules are passed into the blending module to attain more realistic results. Rather than working directly with the RGB texture for inpainting, Grigorev et al.~\cite{Grigorev2018CoordinatebasedTI} predict the coordinates of the texture elements in the input UV-space and extract the colors from the source images to generate the final textures. Finally, Kripasindhu et al.~\cite{Sarkar2020} employ a learned high-dimensional UV feature map to encode the appearance using the inpainting module. Then, the UV feature map is rendered in the desired target pose and is passed through a translation network that creates the final rendered image. However, the results of these methods have visual artifacts as it is difficult for the inpainting model to construct the precise UV texture since it has no UV ground truth for training.

In this paper, we exploit keypoints instead of SMPL which is a skinned vertex-based model, and is a function of shape parameters, pose parameters, and a rigid transformation parameters.

Additionally, some excellent video generation methods have been proposed~\cite{siarohin2019first,liu2021liquid,liu2019liquid,siarohin2021motion,ren2021flow} which aim to transfer the pose from the source video to the target video. We do not refer to these models as our baselines, as our model does not require video sequences as training data.

\begin{figure*}[!ht]\small
\begin{center}
\vspace{-0.2cm}
\includegraphics[width=1.0\linewidth]{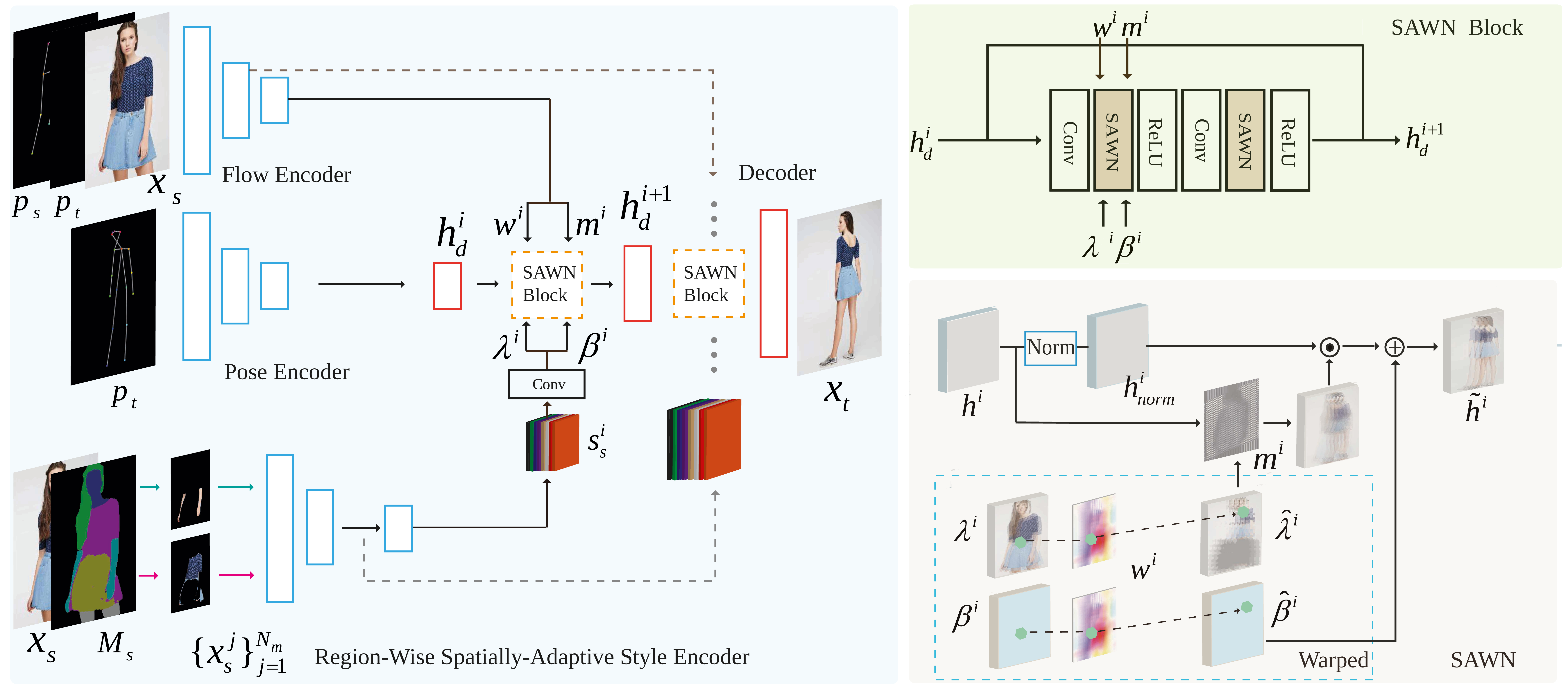}
\end{center}
\vspace{-0.2cm}
\caption{The overview of our architecture with the proposed spatially-adaptive warped normalization~(SAWN). Our architecture consists of one flow encoder, a pose encoder, a region-wise spatially-adaptive style encoder, and one decoder integrating with our SAWN block at multiple scales. Our style encoder attains multi-scale style features $\{s^{i}_{s}\}^{N_{s}}_{i=1}$ where the corresponding channels for style features $s^{i}_{s}$ are attained from the encoder blocks by taking person parts $x^{j}_{s}$ as input. $h^{i}$ represents the features before normalization.}
\label{fig:model}
\end{figure*}

\noindent \textbf{Person Texture-Transfer.} This task aims to produce realistic person images where the texture of one or several person regions is inherited from the reference image, while non-target person regions remain intact. Men et al.~\cite{men2020controllable, men2020controllablepami} proposed a novel architecture~(ADGAN) for this task, and their model regards the pose as content, person image as style, and transfers the style code into the target pose by using a style module based on adaptive instance normalization~\cite{Huang2017ArbitraryST}. Other works~\cite{Zhu2020SemanticallyMI,Zhu_2020_CVPR} are based on the general semantic image synthesis model and can be used to implement the manipulation of the person attributes. However, a large spatial misalignment between source segmentation and target garments is not considered, which causes low-quality images, especially for large pose variations and complex textures. Recently, PISE~\cite{PISE} presented a spatially-aware normalization and used the VGG feature to put the pose and style features into the same feature space. However, PISE also fails in generating and transferring complex textures, as it lacks an explicit spatial warped operation to align pose and style features. Concurrent with our work, Sarkar et al.~\cite{sarkar2021style} proposed a StyleGAN-based method that exploits SMPL~\cite{SMPL:2015} correspondences as pose presentation, and their model achieves high-quality generation and garment transfer results. Compared with it, our model does not exploit this 3D representation information to guide the generation.

Additionally, the person texture-transfer task is very similar to the virtual try-on task, which transfers a desirable clothing item to the corresponding person~\cite{han2017viton,Yu_2019_ICCV,wang2018toward,Yang_2020_CVPR,Issenhuth2020DoNM,10.1145/3450626.3459884,choi2021viton,Zhao2021M3DVTONAM,yang2021ct,Xu_2021_ICCV,ge2021parser,Chopra2021ZFlowGA,ge2021disentangled}. We argue that the differences lie in two aspects. First, our task in this paper is to edit multiple person regions, and the try-on works focus on the garment. Second, in this task, we transfer the complex textures into the source person's garment while preserving the shape of the garment in this source person.

\noindent \textbf{Conditional Normalization} has been widely used for various vision tasks, such as style transfer~\cite{huang2017adain}, image translation~\cite{huang2018munit,choi2020starganv2,park2019SPADE,Zhu_2020_CVPR,CelebAMask-HQ}, super-resolution~\cite{wang2018sftgan} or vanilla generative model~\cite{Karras2019ASG}. Differently from previous unconditional normalization techniques~\cite{ioffe2015batch,ulyanov2016instance,wu2018group}, conditional normalizations~\cite{Dumoulin2017ALR,huang2017adain} require external data that are used to infer the modulation parameters. Then, normalized activations are modulated by the scales and bias parameters previously inferred. Based on adaptive instance normalization (AdaIN)~\cite{huang2017adain}, Taesung et al.~\cite{park2019SPADE} proposed a spatially-adaptive normalization which can effectively propagate the semantic information through the network while preserving the spatial information of the semantic representation. Some variations based on the spatially-adaptive normalization include conditional group normalization by using GroupNet instead of ConvNet~\cite{Zhu2020SemanticallyMI}, semantic region-adaptive normalization~\cite{Zhu_2020_CVPR} with both style and mask inputs for regional image editing, and class-adaptive normalization~\cite{tan2021efficient} which is only applicable to semantic class.

Compared to these previous methods, we integrate the learned flow field into instance normalization, which warps the modulation parameters to align pose and style features.

\begin{figure*}[!ht]\small
\begin{center}
\vspace{-0.2cm}
\includegraphics[width=1.0\linewidth]{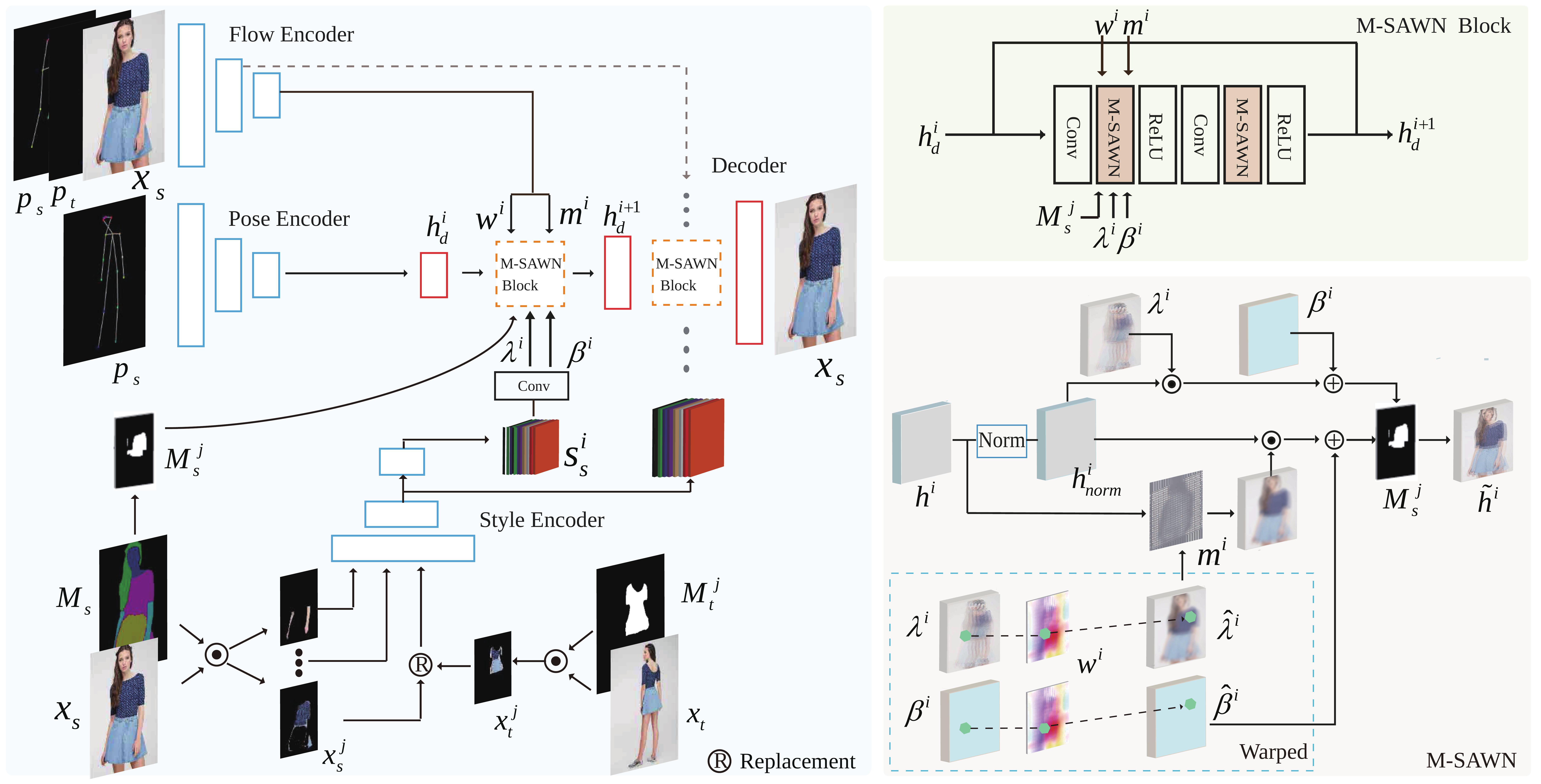}
\end{center}
\vspace{-0.2cm}
\caption{Overview of our proposed Self-Training Part Replacement architecture. This architecture takes the source appearance $x_{s}$ as input of the style encoder and the source pose $p_{s}$ as the input of the pose encoder to reconstruct $x_{s}$. We randomly replace one part $x^{j}_{s}$ using $x^{j}_{t}$ from the target person $x_{t}$ to attain mixed styles. Additionally, the mask $M^{j}_{s}$, which indicates the replacement part, is an input of our M-SAWN module to perform alpha blending.}
\vspace{-0.4cm}
\label{fig:model2}
\end{figure*}

\section{Preliminaries}

Since our method is based on the spatially-invariant adaptive instance normalization~\cite{Huang2017ArbitraryST,Karras2019ASG,men2020controllable} and the spatially-adaptive normalization~\cite{park2019semantic}, we first introduce the key ideas of both approaches.

\noindent \textbf{Spatially-Invariant Adaptive Instance Normalization}~(AdaIN) was initially introduced for the style-transfer task~\cite{Huang2017ArbitraryST}. Let $h^{i} {\in} B {\times} C^{i} {\times} H^{i} {\times} W^{i}$ denote the activations of the $i$-th layer in decoder, where $B$, $C^{i}$, $H^{i}$, and $W^{i}$ represent batch size, number of channels, height, and width, respectively. Spatially-invariant adaptive instance normalization can generally be formulated as:
\begin{equation}
\begin{aligned}
\tilde h^{i}_{b,c,y,x} = \lambda^{i}_{b,c} \frac{h^{i}_{b,c,y,x} - \mu^{i}_{b,c}}{\sigma^{i}_{b,c}}
+ \beta^{i}_{b,c},
\end{aligned}
\end{equation}
where $\tilde h^{i}_{b,c,y,x}$ is the normalized value on the index ($b, c, y, x$), and $b {\in} [0,B{-}1]$, $c {\in} [0,C^{i}{-}1]$, $y {\in} [0,H^{i}{-}1]$, $x {\in} [0,W^{i}{-}1]$. $\lambda^{i} {\in} R^{B \times C^{i}}$ and $\beta^{i} {\in} R^{B \times C^{i}}$ are the learned scale and bias, respectively. Usually, both scale and bias are inferred by a multi-layer perceptron. Finally, $\mu^{i}$ and $\sigma^{i}$ are the mean and standard deviation of input activations $h^{i}$:

\begin{equation}
\begin{aligned}
\mu^{i}_{b,c} & =  \frac{1}{H^{i}W^{i}}\sum_{y,x}h^{i}_{b,c,y,x}, \\
\sigma^{i}_{b,c} & =  \sqrt{ \frac{1}{H^{i}W^{i}}
                 \sum_{y,x}\big( (h^{i}_{b,c,y,x})^{2} - (\mu^{i}_{b,c})^{2} \big)}.
\end{aligned}
\end{equation}

To this end, both the modulation parameters $\lambda^{i}$ and $\beta^{i}$ and the normalization parameters $\mu^{i}$ and $\sigma^{i}$ are the same in spatial coordinates.

\noindent \textbf{Spatially-Adaptive Instance Normalization.}
As mentioned above, the previous baseline ADGAN uses the global style description provided by AdaIN and neglects the person image's spatial information, leading to unrealistic texture transfer. Thus, the modulation parameters $\lambda^{i}$ and $\beta^{i}$ should be spatially-adaptive, i.e., they have the same spatial dimension as the input activations $h^{i}$. Spatially-adaptive instance normalization can be defined as:
\begin{equation}
\begin{aligned}
\tilde h^{i}_{b,c,y,x} = \lambda^{i}_{b,c,y,x} \frac{h^{i}_{b,c,y,x} - \mu^{i}_{b,c}}{\sigma^{i}_{b,c}}
+ \beta^{i}_{b,c,y,x}.
\end{aligned}
\end{equation}

\section{The Proposed Method}

As shown in Fig.~\ref{fig:model}, our architecture consists of a flow encoder, a pose encoder, a style encoder, and a decoder with the spatially-adaptive warped normalization~(SAWN) module.
The pose encoder takes the target pose $p_{t}$ as input and extracts the pose features. The flow encoder takes the source pose $p_{s}$, the target pose $p_{t}$, and the source person image $x_{s}$ as input and outputs multi-scale flow fields $\left\{w^{i}\right\}_{i=1}^{N_{s}}$~(where $N_s$ is the number of scales) and occlusion masks $\left\{m^{i}\right\}_{i=1}^{N_{s}}$. The style encoder takes the source person $x_{s}$ and the corresponding semantic segmentation $M_{s}$ to extract region-aware style codes $s^{i}_{s}$ on multiple scales and then predicts the modulation parameters $\lambda^{i}$ and $\beta^{i}$ using a single conv-block.
The decoder consists of several SAWN blocks taking the learned flow field $w^{i}$, occlusion mask $m^{i}$, modulation parameters $\lambda^{i}$, $\beta^{i}$, and previous layer features $h^{i}_{d}$ as inputs to generate a new feature $h^{i+1}_{d}$. This process is repeated at each scale $i$. Finally, the decoder produces the output image $x_{t}$.  
The SAWN block aims to align the style features with pose features by warping the modulation parameters. Later the warped parameters are used to modulate the normalized features.


\begin{sloppypar}\noindent \textbf{Spatially-Adaptive Warped Normalization.}
Based on the spatially-adaptive instance normalization, we propose a spatially-adaptive warped normalization~(SAWN) to align the modulation parameters $\lambda^{i}$ and $\beta^{i}$ with the target pose feature and then modulate the normalized pose feature using the warped parameters. As shown in the bottom-right of Fig.~\ref{fig:model}, SAWN takes the feature $h^{i}$ and the other four parameters as input: scale $\lambda^{i}$, bias $\beta^{i}$, learned flow-field $w^{i}$, and occlusion mask $m^{i}$. \end{sloppypar}

In detail, we first normalize the features $h^{i}$ to attain normalized features $h^{i}_{norm}$ and then we employ the flow-field $w^{i}$ to warp $\lambda^{i}$ and $\beta^{i}$ to produce the warped parameters $\hat \lambda^{i}$ and $\hat \beta^{i}$ by means of a bilinear sampling:
\begin{equation}
\begin{aligned}
\hat \lambda^{i}(b,c,y,x) & = \lambda^{i}\left[b; c; y + w^{i}_{b,1,y,x}; x + w^{i}_{b,2,y,x}\right], \\
\hat \beta^{i}(b,c,y,x) & = \beta^{i}\left[b; c; y + w^{i}_{b,1,y,x}; x + w^{i}_{b,2,y,x}\right],
\end{aligned}
\end{equation}
where the square brackets represent the bilinear interpolation.
Since the style parameter $\lambda^{i}$ inferred from the source appearance does not provide all the content of the target appearance, due to the frequent self-occlusion, we perform an alpha blending between the scale $\hat \lambda^{i}$ and input activations $h^{i}$ using the occlusion mask $m^{i}$. Therefore, the proposed SAWN can be defined as:
\begin{equation}
\tilde h^{i} = \big((\hat\lambda^{i} \odot m^{i} + h^{i} \odot (1 - m^{i}) \big) \odot h^{i}_{norm} \oplus \hat \beta^{i}.
\end{equation}
As shown in Fig.~\ref{fig:model}, we use SAWN in every scale of the decoder. We tested several blending combinations and found that blending $h^{i}$ and $\hat \lambda^{i}$ performs best. Additionally, we do not perform the blending for the parameter $\hat \beta^{i}$, as the style information is predominantly provided by the scale and not by the bias~\cite{karras2020analyzing}.

\noindent \textbf{Region-Wise Spatially-Adaptive Encoder.} We employ semantic segmentation to disentangle person attributes, such as garment and hair, to perform texture transfer guided by the reference person region. As shown in the bottom-left of Fig.~\ref{fig:model}, we obtain the person parts $\{x^{j}_{s}\}^{N_{m}}_{j=1}$~($N_{m}$ is the number of segmentation labels) using the corresponding segmentation mask $M_{s}^{j}$ multiplied 
by the person image $x_{s}$. Then, each part of $\{x^{j}_{s}\}^{N_{m}}_{j=1}$ is used as input of the style encoder to obtain the corresponding style codes. Finally, the style codes for all parts are concatenated into $s^{i}_{s}$, which is later processed by a single conv-block to produce the spatially-adaptive modulation parameters $\lambda^{i}$ and $\beta^{i}$. Since the style codes are separately extracted from each person part and then concatenated, the input of the conv-block is disentangled with respect to the specific person regions, which facilitates replacing the individual parts of the source person with the corresponding target parts.
We refer to our style encoder as the region-wise spatially-adaptive style encoder.

\noindent \textbf{Self-Training Part Replacement Strategy and M-SAWN.}
The previous texture-transfer methods~\cite{men2020controllable,PISE} are usually trained in the same way as pose-guided generation methods. At the test stage, the style code of the source person region is replaced with the reference person region. Since the model never observed such mixed style codes from different images during training, this causes a substantial-quality degradation and bad texture preservation in the generated images. Thus, we propose a self-training part replacement architecture that integrates the part replacement operation with a self-training strategy.

\begin{figure*}[!t]\small
\begin{center}
\includegraphics[width=1\linewidth]{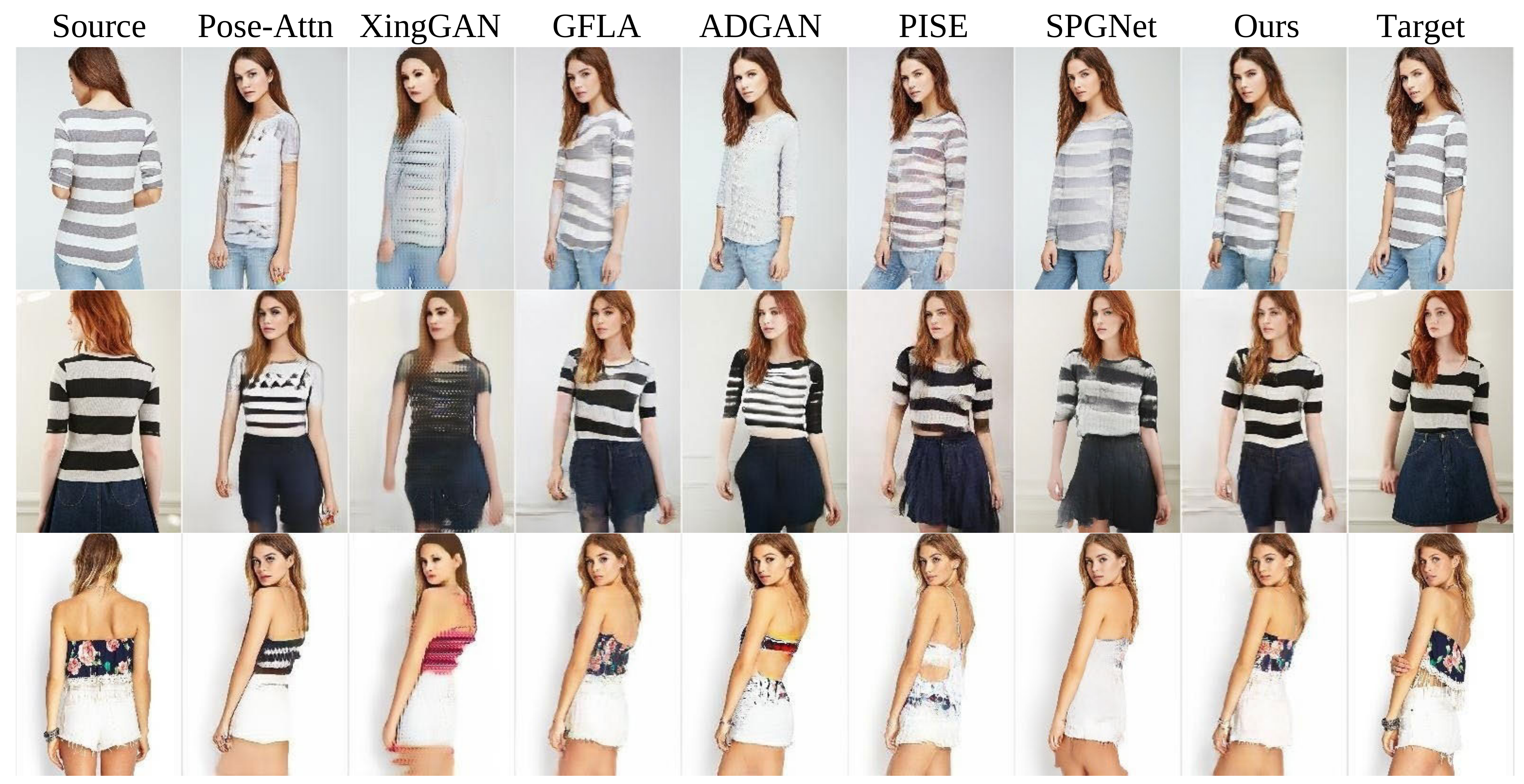}
\end{center}
\vspace{-0.2cm}
\caption{Qualitative comparison between our method and the state-of-the-art methods, i.e., Pose-Attn~\cite{zhu2019progressive}, XingGAN~\cite{tang2020xinggan}, GFLA~\cite{ren2020deep}, ADGAN~\cite{men2020controllable}, PISE~\cite{PISE} and SPGNet~\cite{lv2021learning} on the DeepFashion dataset.}
\vspace{-0.2cm}
\label{fig:exp1}
\end{figure*}

\begin{figure*}[!t]\small
\begin{center}
\vspace{-0.2cm}
\includegraphics[width=0.98\linewidth]{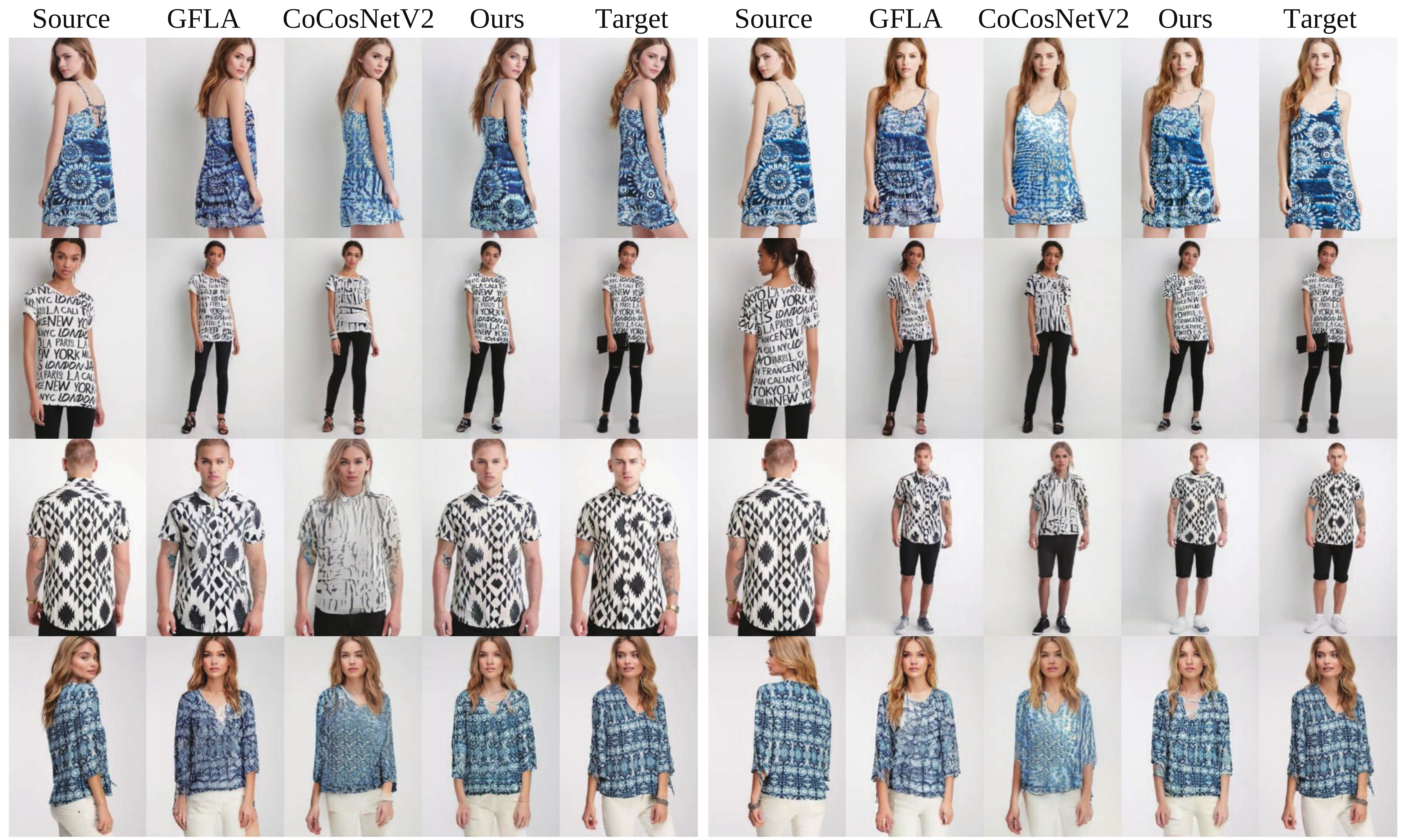}
\end{center}
\vspace{-0.3cm}
\caption{Qualitative comparison between ours~(256) and the GFLA (256)~\cite{ren2020deep}, CocosNetV2 (256)~\cite{Zhou_2021_CVPR} on high-resolution DeepFashion.}
\vspace{-0.5cm}
\label{fig:exp12}
\end{figure*}

\begin{table*}[!t] \small
\caption{Quantitative results between ours and state-of-the-art methods on DeepFashion. GFLA~(256)~\cite{ren2020deep}, CoCosNetV2~(256)~\cite{Zhou_2021_CVPR}, and Ours~(256) indicates both methods are trained with $256{\times} 256$ deepfashion images~(resized from $750 {\times}  1101 $).}
\centering
\resizebox{0.8\linewidth}{!}{%
\begin{tabular}{rccccc} \toprule
Method &  FID$\downarrow$  & LPIPS$\downarrow$ & LFID$\downarrow$  & LLPIPS$\downarrow$  & User Study$\uparrow$ \\ \midrule
Pose-Attn~\cite{zhu2019progressive} & 20.739 & 0.253 & 47.56 & 0.287 & 5.50\% \\
XingGAN~\cite{tang2020xinggan} & 39.233 & 0.282 & 73.51 & 0.333 & 1.350\% \\

GFLA~\cite{ren2020deep} & 15.573 & 0.232 & 38.36 & 0.247 & 26.35\% \\
ADGAN~\cite{men2020controllable} & 16.000 & 0.224 & 37.01 & 0.262 & 8.18\% \\
PISE~\cite{PISE} &  13.610 & 0.205 & 36.46 & 0.240 & 14.70\% \\
SPGNet~\cite{lv2021learning} & 12.243 & 0.210 & 37.42 & 0.259 & 13.70\% \\
\hline
Ours & \textbf{11.540} & \textbf{0.196} & \textbf{34.22} & \textbf{0.229} & \textbf{30.22\%} \\
\bottomrule
GFLA~(256) \cite{ren2020deep} & \textbf{10.573} & 0.234 & 14.01 & 0.246 & 30.25\% \\
CoCosNetV2~(256)~\cite{Zhou_2021_CVPR} & 13.900 & 0.228 & 13.94 & 0.267 &  21.15\% \\
Ours~(256) & 11.210 & \textbf{0.219} & \textbf{11.82} & \textbf{0.242} &  \textbf{48.60\%} \\
\bottomrule
\end{tabular}}
\label{tab:quan1}
\end{table*}

In detail, we use the same training data as the architecture of Fig.~\ref{fig:model}, but there are two important differences. As shown in Fig.~\ref{fig:model2}, we use the source pose $p_{s}$ and person $x_{s}$ to reconstruct $x_{s}$ instead of target person $ x_{t}$, and we will randomly replace one part $x_{s}^{j}$ of the source person $x_{s}$ using part $x_{t}^{j}$ from the target person $x_{t}$ to produce mixed style codes in our style encoder. Second, we propose to use the segmentation mask $M^{j}_{s}$ from $M_{s}$ as an additional input of the SAWN block, which indicates the part that needs to be replaced (M-SAWN). Note that $M^{j}_{s}$ is used to perform local warped modulation and to preserve other regions. Furthermore, we use the same flow encoder as in the architecture of Fig.~\ref{fig:model}. 
As shown in Fig.~\ref{fig:model2}, the M-SAWN is defined as:
\begin{equation} \label{gcn1}
\begin{aligned}
\tilde h^{i}_{w} & = \big((\hat\lambda^{i} \odot m^{i} + h^{i} \odot (1 - m^{i}) \big) \odot h^{i}_{norm} \oplus \hat \beta^{i}, \\
\tilde h^{i}_{nw} & = (\lambda^{i} \odot h^{i}_{norm}) \oplus \beta^{i}, \\
\tilde h^{i} & = \tilde h^{i}_{w} \odot M^{j}_{s} + \tilde h^{i}_{nw} \odot (1 - M^{j}_{s}),
\end{aligned}
\end{equation}
where $\tilde h^{i}_{w}$ and $\tilde h^{i}_{nw}$ indicate the modulation features obtained using the warped parameters and obtained without the warped parameters, respectively.


\noindent \textbf{Training Losses.}
Similar to the previous baselines~\cite{PISE,men2020controllable}, we employ four losses: adversarial loss $\mathcal{L}_{adv}$, $L1$ reconstruction loss $\mathcal{L}_{recon}$, VGG style loss $\mathcal{L}_{vgg_s}$, and VGG content loss $\mathcal{L}_{vgg_c}$, for the entire training. The overall loss can be defined as:
\begin{eqnarray}
\mathcal{L}_{all} =\lambda_{1} \mathcal{L}_{adv} + \lambda_{2} \mathcal{L}_{recon} + \lambda_{3} \mathcal{L}_{vgg_s} + \lambda_{4} \mathcal{L}_{vgg_c},
\label{eq:overall1}
\end{eqnarray}
where $\lambda_{1}$, $\lambda_{2}$, $\lambda_{3}$, and $\lambda_{4}$ are hyperparameters that control the contribution of each loss term. 


\subsection{Differences from Related Methods}

 There are several key differences between our method and other flow-based models, such as FOMM~\cite{siarohin2019first} and GFLA~\cite{ren2020deep}. First, we apply the learned flow fields to warp the parameters of the modulation operations instead of the decoder features. It allows our model to handle both pose-transfer and texture-transfer tasks seamlessly. Note that FOMM and GFLA cannot deal with texture-transfer tasks. Second, M-SAWN could be used to solve the local misalignment problem for a region-specific texture transfer, while FOMM and GFLA could not handle this task. On the other hand, existing methods for this task, e.g., SEAN~\cite{Zhu_2020_CVPR}, suffer from the misalignment problem between semantics and textures. Finally, modulation parameters (scales and biases) denormalize the features with different operations: Hadamard product and matrix addition. It provides required flexibility and scalability for the model, which is beneficial for learning the alignment of complex textures (see Section~\ref{ablation}).

\section{Experiments}

\noindent\textbf{Datasets.} We evaluate our model on the DeepFashion~\cite{liu2016deepfashion} (in-shop clothes retrieval benchmark) dataset, which is widely used in person image generation tasks.
DeepFashion contains 52,712 person images with various poses and appearances and has low-resolution~($176{\times} 256$) and high-resolution images~($750{\times} 1101$). We resize all the images into $256 {\times} 256$ for both training and testing. We split the dataset following ADGAN~\cite{men2020controllable}; other state-of-the-art methods also adopt the same data configuration. In more detail, 101,966 pairs are selected for training and 8,570 for testing. Furthermore, we use the same segmentation masks for person images as in ADGAN, obtained from the human parser~\cite{Badrinarayanan2017SegNetAD}. We sample 8,570 pairs with different identities from the test dataset to evaluate the texture transfer task and select each pair's `Tops', `Pants', `Hair' regions.

\noindent\textbf{Implementation Details.} 
To obtain accurate flow field estimation, we use the local-attention module of GFLA~\cite{ren2020deep} instead of the original bilinear sampling to avoid poor gradient propagation. Moreover, we pre-train the flow encoder using their sampling correctness loss and flow regularization loss. We set $\lambda_{1}{=}2.0$, $\lambda_{2}{=}5.0$, $\lambda_{3}{=}0.5$, and $\lambda_{4}{=}0.0025$. Furthermore, we use the Instance Normalization~\cite{Huang2017ArbitraryST} for the layers in both the discriminator and generator. We use the ADAM~\cite{kingma2014adam} optimizer with $\beta_{1}{=}0.5$, $\beta_{2}{=}0.999$, and learning rate $0.0001$ for training our model. All of the experiments are conducted on two 16GB Tesla V100 GPUs with the PyTorch framework~\cite{paszke2019pytorch}. Note that our finetune only needs 100 steps with 10 minutes.

\begin{figure*}[!ht]\small
\begin{center}
\vspace{-0.1cm}
\includegraphics[width=1.0\linewidth]{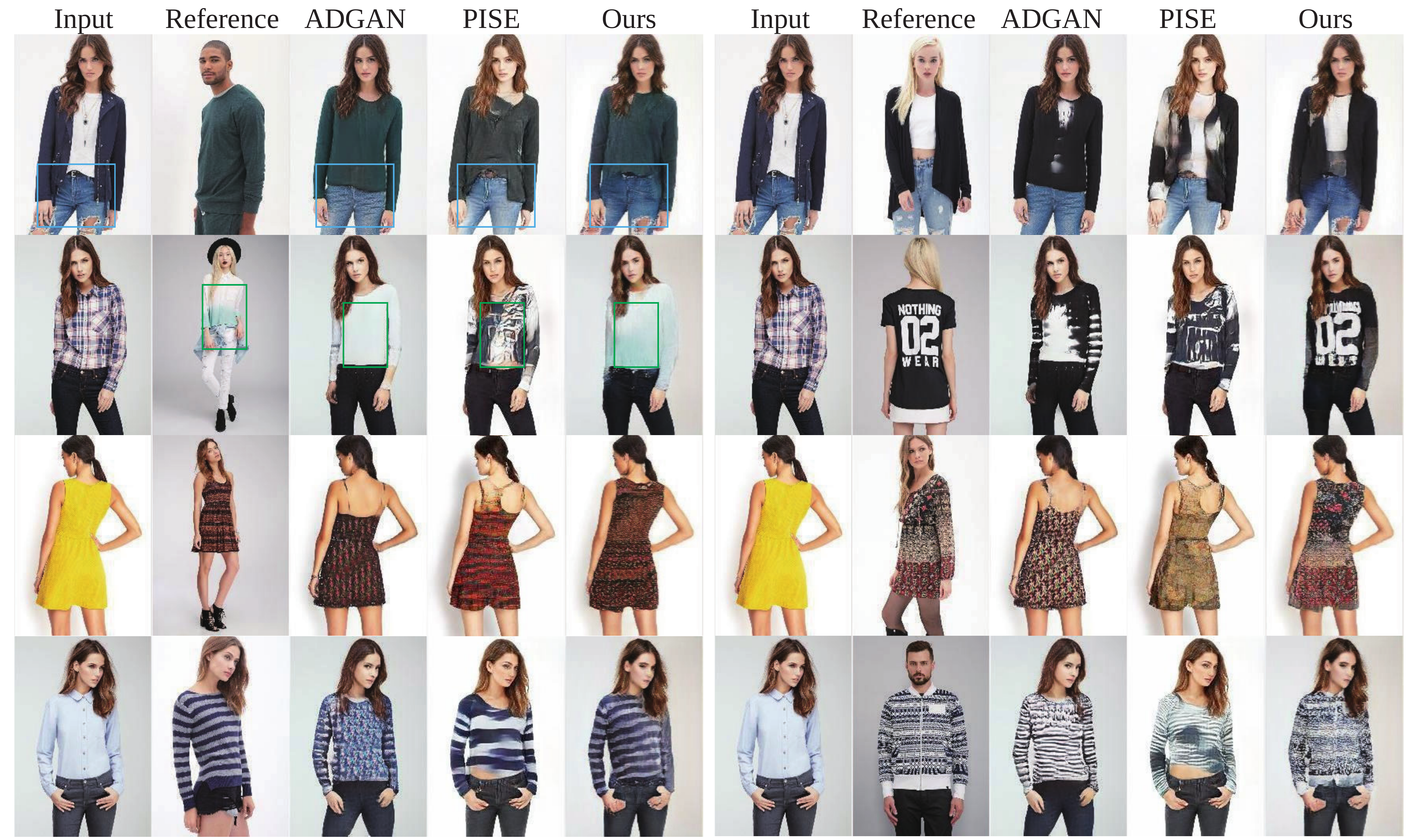}
\end{center}
\vspace{-0.3cm}
\caption{Qualitative comparison of texture transfer between our method, ADGAN~\cite{men2020controllable}, and PISE~\cite{PISE}. Blue box: comparison of non-target attribute preservation. Green Box: comparison of garment texture transfer.}
\vspace{-0.4cm}
\label{fig:exp2}
\end{figure*}

\begin{figure*}[!t]\small
\begin{center}
\includegraphics[width=1.0\linewidth]{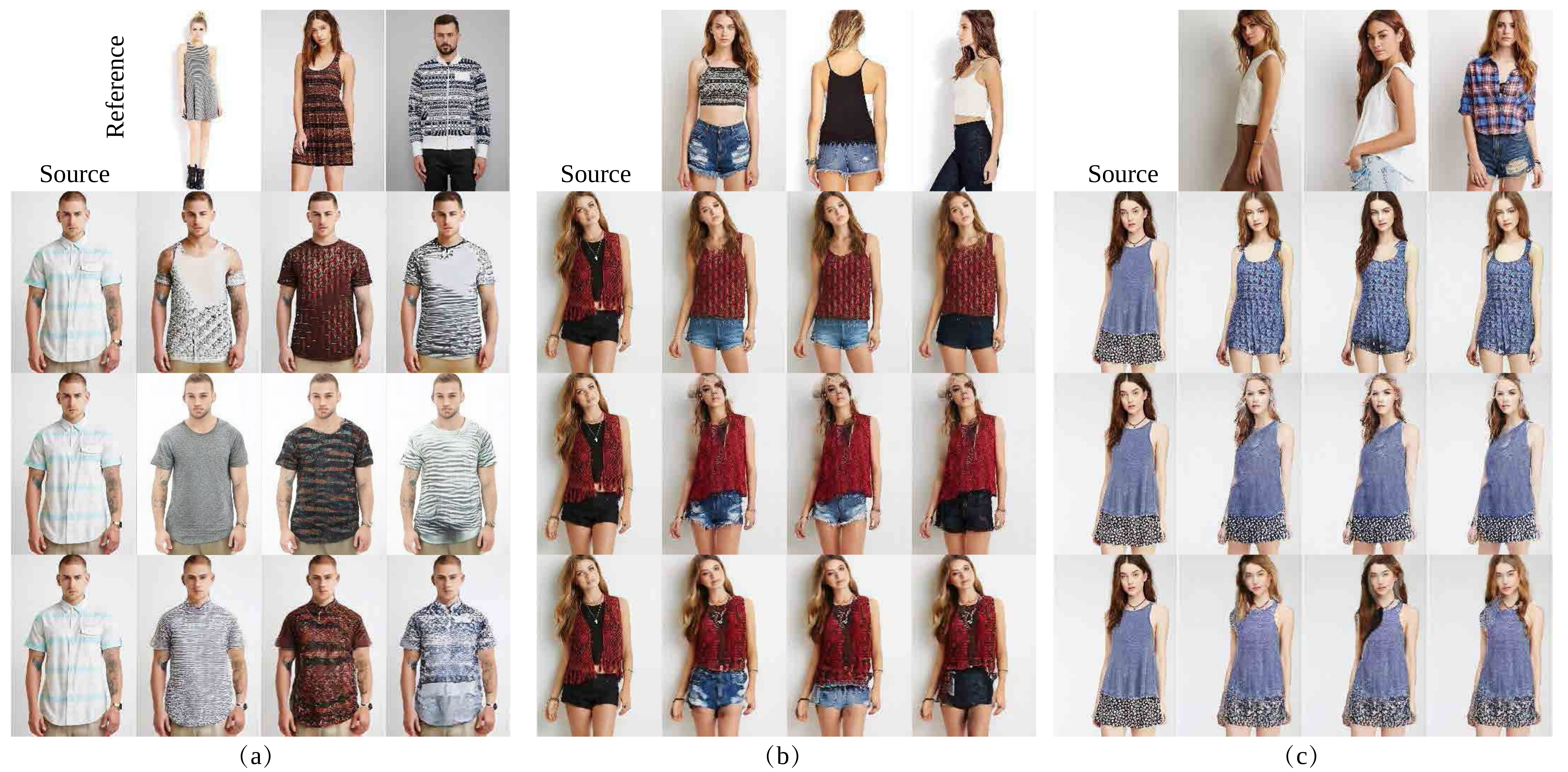}
\end{center}
\vspace{-0.3cm}
\caption{Qualitative comparison of texture transfer between our method (4th row), ADGAN~\cite{men2020controllable} (2nd row), and PISE~\cite{PISE} (3rd row). (a, b, c) represent tops, pants, and hair transfer, respectively.}
\vspace{-0.5cm}
\label{fig:exp22}
\end{figure*}

\begin{figure*}[t]\small
\begin{center}
\includegraphics[width=1.0\linewidth]{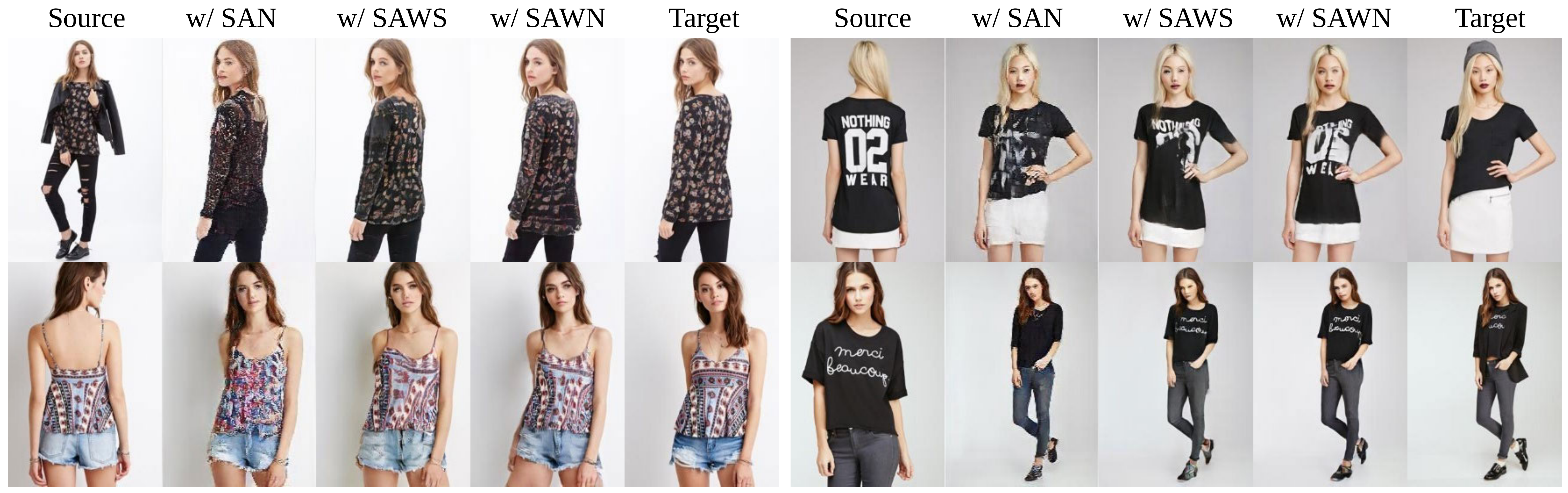}
\end{center}
\vspace{-0.3cm}
\caption{Comparison between our full model~(with SAWN) and two variants: SAN, SAWS. SAN is the original spatially-adaptive instance normalization without warping operation. SAWS is the spatially-adaptive instance normalization with warping scales, and without warping bias.}
\label{fig:ab}
\vspace{-0.3cm}
\end{figure*}

\begin{figure*}[t]\small
\begin{center}
\vspace{-0.1cm}
\includegraphics[width=1.0\linewidth]{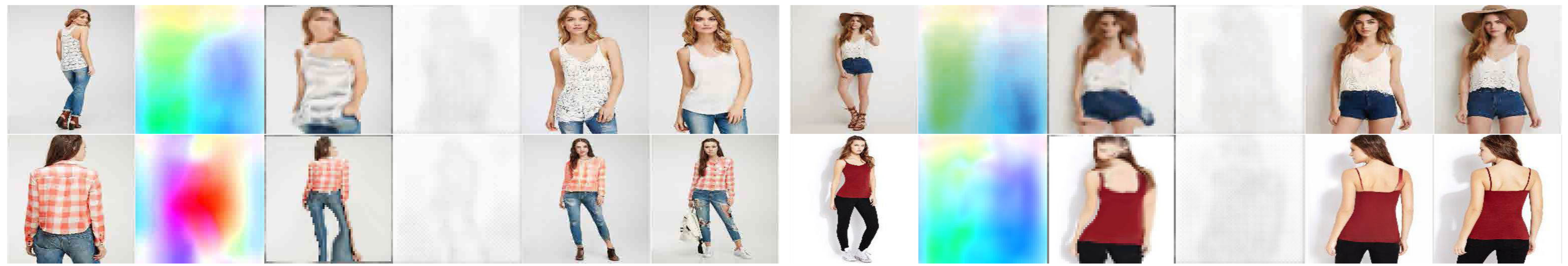}
\end{center}
\vspace{-0.2cm}
\caption{Visualization of the learned flow field and occlusion mask. 1st column is the input person, and the following columns are learned flow fields, the warped results using the flow fields for input images, occlusion masks, generated results, and target images.}
\label{fig:ab3}
\vspace{-0.3cm}
\end{figure*}

\begin{figure*}[t]\small
\begin{center}
\includegraphics[width=1.0\linewidth]{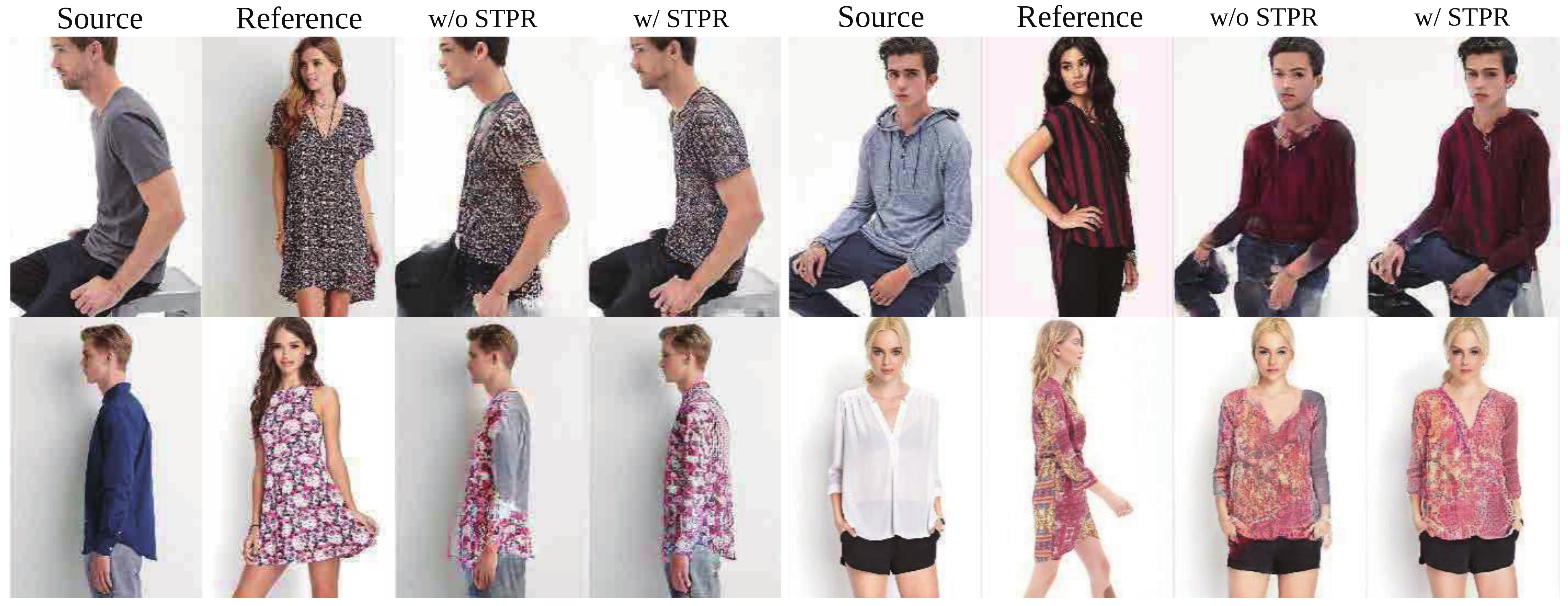}
\end{center}
\vspace{-0.3cm}
\caption{Comparison between our full model and it without self-training part replacement strategy~(STPR).}
\label{fig:ab2}
\vspace{-0.3cm}
\end{figure*}

\begin{table*}[t]
\scriptsize
\caption{Quantitative results between ours with state-of-the-art methods on DeepFashion.}
	\centering
    \begin{threeparttable}
    	\resizebox{0.7\linewidth}{!}{
		\begin{tabular}{rcccccc} \toprule
			\multirow{2}{*}{Method} & \multicolumn{2}{c}{Tops} & \multicolumn{2}{c}{Pants} & \multicolumn{2}{c}{Hair}  \\ \cmidrule(lr){2-3} \cmidrule(lr){4-5} \cmidrule(lr){6-7} 
			 & FID$\downarrow$ & MLPIPS$\downarrow$ & FID $\downarrow$ & MLPIPS $\downarrow$  & FID $\downarrow$  & MLPIPS$\downarrow$ \\ \midrule	
            ADGAN & 14.37 & 0.195 & 14.80 & 0.057 & 15.23 & 0.061  \\
            PISE & 18.55 & 0.190  & 18.95  & 0.056 & 18.84 & \textbf{0.057}  \\
            \hline
            Ours & \textbf{11.92} & \textbf{0.056}  & \textbf{10.50}  & \textbf{0.055} & \textbf{11.32} & 0.061 \\
			\bottomrule	
	\end{tabular}}
	\end{threeparttable}
    \label{tab:quan2}
    \vspace{-0.3cm}
\end{table*}

\noindent\textbf{Evaluation Metrics.} 
The selection of metrics to assess the quality of the generated images remains an open problem. The previous methods~\cite{siarohin2018deformable,zhu2019progressive} exploit Inception score~(IS)~\cite{salimans2016improved} to evaluate the quality of the generated samples and SSIM~\cite{wang2004image} to evaluate the similarity between the generated samples and the ground truth. Recent works have verified that Fr\'echet Inception Distance~(FID)~\cite{NIPS2017_7240} and Learned Perceptual Image Patch Similarity~(LPIPS)~\cite{zhang2018perceptual} are more correlated with human evaluation than Inception score~\cite{salimans2016improved} and SSIM~\cite{wang2004image}. Thus, we follow GFLA~\cite{ren2020deep} and utilize FID and LPIPS to evaluate all models for pose transfer. Furthermore, we crop the local region to evaluate the quality of complex textures and refer to FID and LPIPS for the local region as LFID and LLPIPS, respectively.

Additionally, for the evaluation of region-specific texture transfer, we employ FID to measure the realism of the generated images and exploit LPIPS to evaluate the consistency of the non-target regions that should not be transferred, which can be obtained by using the corresponding mask. We refer to LPIPS for non-target regions as MLPIPS. Lastly, we conduct user studies to assess the subjective quality and ask volunteers to select the most realistic image from the ground truth and generated images. Specifically, 20 volunteers were asked to choose the most realistic synthesized image from all models. Each of them was asked 50 questions. 

\subsection{State-of-the-Art Comparison}

\noindent\textbf{Pose Transfer.} \label{exp1}
The qualitative comparisons with state-of-the-art methods are shown in Fig.~\ref{fig:exp1}. We can observe that Pose-Attn, XingGAN, and ADGAN have serious mode collapse appearance problems. GFLA, PISE, and SPGNet, on the other hand, suffer from serious appearance inconsistencies when generation assumes large spatial deformation. In contrast, our model achieves sharper results with better appearance consistency, especially for complex texture transfer, such as the white and grey t-shirt in the 1st row. Moreover, Fig.~\ref{fig:exp12} shows that our model (256) can attain better preservation of the complex texture, but GFLA(256) and CoCosNetV2 (256) suffer from unrealistic distortions of texture. The quantitative results are shown in Table~\ref{tab:quan1} that highlight the improvement of our model in this task. For low-resolution results, our model outperforms other methods in all metrics, demonstrating that our results are more realistic and have better consistency in appearance and pose. Our model achieves a close FID score for the high-resolution results and dramatically improves the LPIPS score from 0.234 to 0.219 than GFLA (256).
Additionally, our model achieves better LFID and LLPIPS scores than other models, confirming that our model can generate a more realistic and consistent texture. Moreover, the user study shows that our model generates the most realistic results. More results can be found in the demo video provided.


\noindent\textbf{Texture Transfer.} \label{exp2} 
We provide texture-transfer results guided by different reference person regions in Fig.~\ref{fig:exp2} and Fig.~\ref{fig:exp22}. Fig.~\ref{fig:exp2} shows that our model achieves a more realistic texture transfer~(see green box). Given reference person images with complex textures in garments, our model can render this texture into the garment region of the source image while preserving other regions, such as jeans~(see blue box). In detail, ADGAN suffers from mode collapse in texture generation and fails to generate sharp images. Meanwhile, PISE achieves sharper results than ADGAN but worse than our model. Moreover, it is hard for PISE to preserve other regions, such as the face and hair. We show more transfer results in different regions, such as pants, and hairstyle, in Fig.~\ref{fig:exp22}. We arrive at a similar conclusion when comparing ours with ADGAN and PISE. 
Quantitative results are shown in Table~\ref{tab:quan2}. Our model achieves better FID and MLPIPS scores in the `Tops' and `Pants' regions and better FID scores in all semantic regions. It indicates our model generates more realistic person images and preserves better consistency of non-target attributes for `Pants' and `Tops' transfer tasks. More texture-transfer results can be found in the provided demo video.

\begin{table}[t] 
\caption{Quantitative results between the full model~(with SAWN) and two variants: SAN, SAWS. SAN is the original spatially-adaptive instance normalization without warping operation. SAWS is the spatially-adaptive instance normalization with warping scales and without warping bias.}
\centering
\normalsize
\resizebox{0.5\linewidth}{!}{%
\begin{tabular}{lcc} \toprule
Method &  FID$\downarrow$  & LPIPS$\downarrow$  \\ \midrule
w/ SAN &  14.99 & 0.207 \\
w/ SAWS &  12.04 & 0.204 \\
w/ SAWN & \textbf{11.54} & \textbf{0.196} \\
\bottomrule
\end{tabular}}

\label{tab:ab1}
\vspace{-0.3cm}
\end{table}

\subsection{Ablation Study} \label{ablation}

\noindent\textbf{Effect of SAWN.} We ablate over two different parts of Sawn to explore the effect on the final generated results. The first variant is normalization without the warping module and is referred to as SAN~(spatially-adaptive normalization without the warping module); the second variant is normalization with warping the modulation scales and is referred to as SAWS (spatially-adaptive normalization with warping scales). We present in Fig.~\ref{fig:ab} the comparison between our full model with these two variants. Compared with SAN, our full model achieves sharper results and can transfer the complex texture from the source image into the target pose. Specifically, Fig.~\ref{fig:ab} shows that the full model could transfer the red-and-white texture to the target, while the variant with SAN could not. Table~\ref{tab:ab1} shows that our full model outperforms the variant with SAN on both FID and LPIPS metrics, which further demonstrates the effectiveness of our proposed SAWN. Compared with SAWS, Fig.~\ref{fig:ab} shows that both SAWS and SAWN achieve similar results, but SAWN performs better than SAWS in the preservation of the complex textures, which validates the necessity of warping both scales and the bias. The scores in Table~\ref{tab:ab1} also quantitatively demonstrate the improvement of SAWN. 

We also visualize the learned flow fields and occlusion masks in Fig.~\ref{fig:ab3}. We apply the learned flow fields~(2nd column) to directly warp the input person~(1st column) and obtain the warped result in pixel level~(3rd column). It can be seen that the warped result at the pixel level has a very similar shape and appearance as the generated result~(5th column) and target person~(6th column), which further validates the ability of our model to learn reasonable flow fields. Additionally, the learned occlusion mask~(4th column) has the same shape as the target person, which indicates that our model can focus on the foreground region and select the most helpful information from the source image.

\begin{table*}[t] \small
\caption{Quantitative results between our full model and the model without STPR. STPR: self-training part replacement strategy.}
\scriptsize
	\centering
    \begin{threeparttable}
    \resizebox{0.7\linewidth}{!}{
		\begin{tabular}{rccccccc} \toprule
			\multirow{2}{*}{Method} & \multicolumn{2}{c}{Tops}  & \multicolumn{2}{c}{Pants} & \multicolumn{2}{c}{Hair}  \\ \cmidrule(lr){2-3} \cmidrule(lr){4-5} \cmidrule(lr){6-7}
			 & FID$\downarrow$ & MLPIPS$\downarrow$ & FID $\downarrow$ & MLPIPS $\downarrow$  & FID $\downarrow$ & MLPIPS $\downarrow$ \\ \midrule
            w/o STPR & 15.72 & 0.057 & 12.13  & 0.056 & 13.00 & \textbf{0.061} \\
            \hline
            Our full model & \textbf{10.50} & \textbf{0.054} & \textbf{0.056} & \textbf{0.055} & \textbf{11.32} & \textbf{0.061}  \\
			\bottomrule	
	\end{tabular}}
	\end{threeparttable}

	\label{tab:ab2}
	\vspace{-0.4cm}
\end{table*}

\noindent \textbf{Effect of STPR.}
Fig.~\ref{fig:ab2} shows that our full model could transfer the complex texture into the target person while preserving the pose of the source and face identity. However, the variant without STPR introduces substantial-quality degradation in the face and garment.
Table~\ref{tab:ab2} presents the quantitative comparison. Specifically, the full model with STPR shows obvious improvements in FID scores over all regions, which shows that STPR improves the realism of the transferred results. Moreover, the full model with STPR attains better MLPIPS scores in most regions, demonstrating better preservation of the non-target source parts.

\section{Limitations and Conclusion}

\noindent \textbf{Limitations.} The results of our model are not always positive. Fig.~\ref{fig:exp2} (4th row) shows that some parts of the human image from our model are not realistic, such as hands. We can explain that it is due to the inaccuracy of both the learned flow field and the predicted segmentation mask. Therefore, exploring the learning of the flow field and improving human parsing will be in future works.

\noindent \textbf{Conclusion.} We propose a novel spatially-adaptive warped normalization (SAWN) that warps the modulation parameters to align the style and pose features. This block significantly improves the quality of the generated person image with complex poses and achieves very realistic texture transfer, working particularly well for complex textures. Moreover, we present a self-training part replacement strategy that further improves the quality of generated images while preserving the identity of non-target regions.

\bibliographystyle{IEEEtran}
\bibliography{sawn}


 




\begin{IEEEbiography}[{\includegraphics[width=1in,height=1.25in,clip]{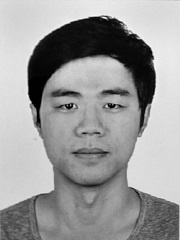}}]{Jichao Zhang} received the M.S. degree from School of Computer Science and Technology, Shandong University in 2019. He is currently pursing the Ph.D. degree from Multimedia and Human Understanding Group at the University of Trento. His research interests include deep generative model, 2D and 3D image generation and editing.
\end{IEEEbiography}

\begin{IEEEbiography}[{\includegraphics[width=1in,height=1in,clip]{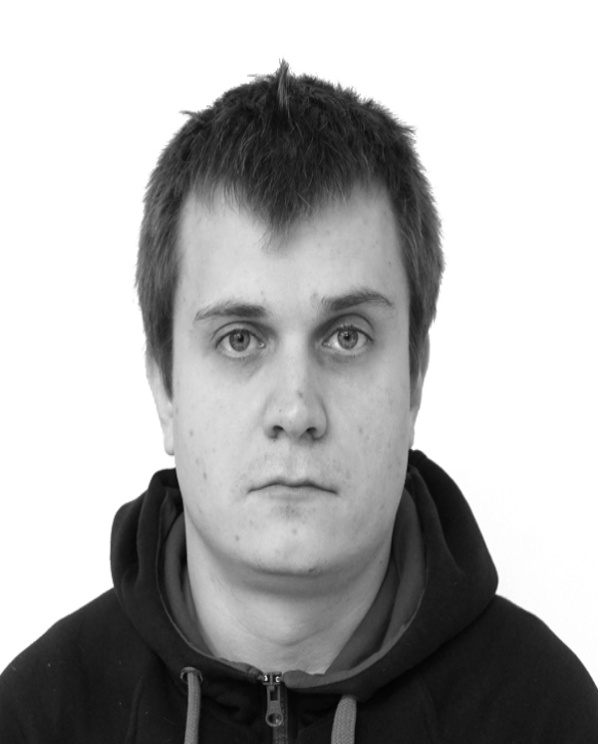}}]{Aliaksandr Siarohin} is a Research Scientist at Snap Research. He received the PhD degree from the Multimedia and
Human Understanding Group, University of Trento,
Italy. His research interests include machine learning for image animation, video generation, generative adversarial networks and domain adaptation.
\end{IEEEbiography}

\vspace{0.0cm}
\begin{IEEEbiography}[{\includegraphics[width=1in,height=1.25in,clip]{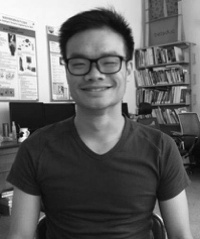}}]{Hao Tang} is currently a Postdoctoral researcher with Computer
Vision Lab, ETH Zurich, Switzerland. He received
the master’s degree from the School of Electronics and Computer Engineering, Peking University,
China and the Ph.D. degree from the Multimedia and
Human Understanding Group, University of Trento,
Italy. He was a visiting scholar in the Department
of Engineering Science at the University of Oxford.
His research interests are deep learning, machine
learning, and their applications to computer vision.
\end{IEEEbiography}

\vspace{0.0cm}

\begin{IEEEbiography}[{\includegraphics[width=1in,height=1.25in,clip]{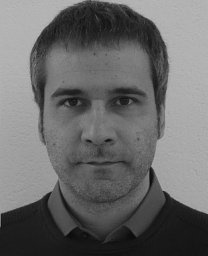}}]{Enver Sangineto} is an Assistant Professor with University of Modena and Regio Emilia, Italy. 
He received his PhD in Computer Engineering from the University of Rome ``La Sapienza''. He has been a post-doctoral researcher at 
the Universities of Rome ``Roma Tre'' and ``La Sapienza'' and at the Italian Institute of Technology (IIT) in Genova.
 His research interests include both discriminative and generative methods and 
learning with minimal human supervision.
\end{IEEEbiography}

\vspace{0.0cm}

\begin{IEEEbiography}[{\includegraphics[width=1in,height=1.25in,clip]{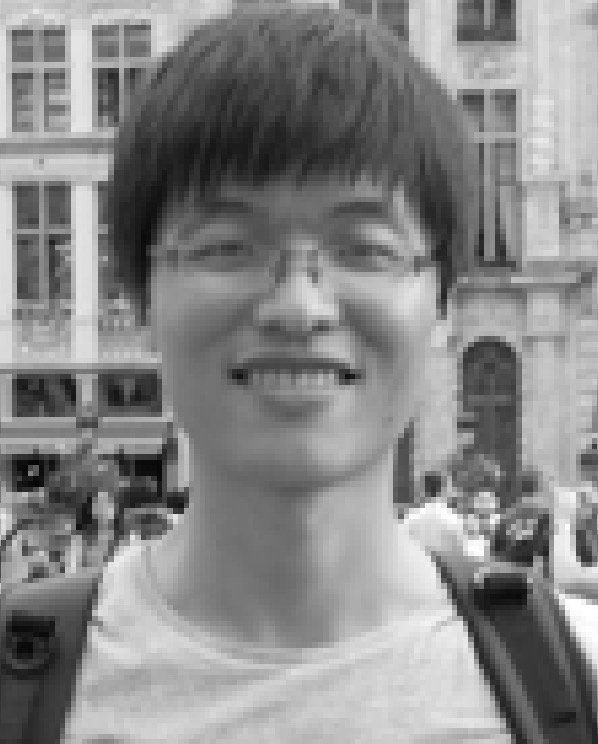}}]{Wei Wang} is currently an Assistant Professor in the Department of Information Engineering and Computer Science at University of Trento, Italy in which he received his Ph.D Degree. He was a Postdoctoral Research Fellow in CVLab at EPFL. His research interests include face analysis, human action understanding, augmented reality (AR), segmentation, optimization problems, etc.
\end{IEEEbiography}

\vspace{0.0cm}

\begin{IEEEbiography}[{\includegraphics[width=1in,height=1in,clip]{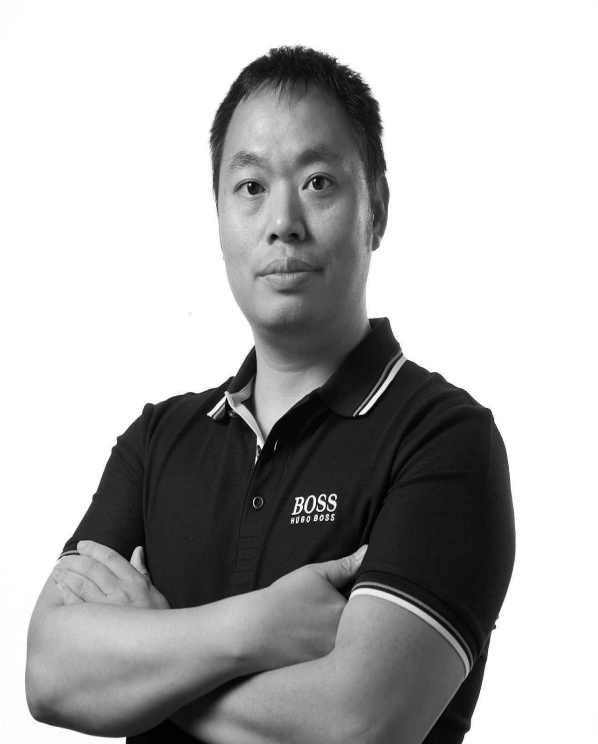}}]{Humphrey Shi} is currently an Assistant Professor of Computer Science, University of Oregon. His recent research focuses on accurate and efficient visual understanding and deep learning
for intelligent systems and applications under various learning settings and input qualities. He was the Area Chair of CVPR 2021, 2022, ICCV 2021, ECCV 2020, 2022.
\end{IEEEbiography}

\vspace{0.0cm}

\begin{IEEEbiography}[{\includegraphics[width=1in,height=1.25in,clip]{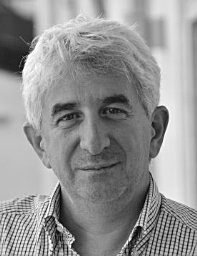}}]{Nicu Sebe} is Professor with the University of Trento, Italy, leading the Multimedia and Human Understanding Group. He was the General Co-Chair of ACM Multimedia 2013 and 2022, and the Program Chair of  ACM Multimedia 2007 and 2011, ICCV 2017, ECCV 2016, ICPR 2020 and ECCV 2024. He is a fellow of the International Association for Pattern Recognition.
\end{IEEEbiography}

\vfill

\end{document}